\UseRawInputEncoding

\documentclass[10pt,twocolumn,letterpaper]{article}

\usepackage{cvpr}              

\usepackage{graphicx}
\usepackage{bbding}
\usepackage{amsmath}
\usepackage{amssymb}
\usepackage{booktabs}
\usepackage{amsmath}
\usepackage{comment}
\usepackage[titlenumbered,ruled]{algorithm2e}
\usepackage[pagebackref,breaklinks,colorlinks]{hyperref}

\usepackage[capitalize]{cleveref}
\crefname{section}{Sec.}{Secs.}
\Crefname{section}{Section}{Sections}
\Crefname{table}{Table}{Tables}
\crefname{table}{Tab.}{Tabs.}

\def\cvprPaperID{3357} 
\def\confName{CVPR}
\def\confYear{2022}

\begin{document}

\title{Few-Shot Font Generation by Learning Fine-Grained Local Styles}
\author{
Licheng Tang\textsuperscript{1}\thanks{Equal contribution.} 
\quad
Yiyang Cai\textsuperscript{2}$^*$ 
\quad
Jiaming Liu\textsuperscript{1}\thanks{Corresponding author.}, 
\quad
Zhibin Hong\textsuperscript{1} 
\quad
Mingming Gong\textsuperscript{3} \\
\quad 
Minhu Fan\textsuperscript{1} \quad
Junyu Han\textsuperscript{1} \quad
Jingtuo Liu\textsuperscript{1} \quad
Errui Ding\textsuperscript{1} \quad
Jingdong Wang\textsuperscript{1}
\\ 
 \textsuperscript{1}Baidu Inc.
\quad
 \textsuperscript{2}University of California, Berkeley 
 \quad
 \textsuperscript{3}University of Melbourne \\
  {\tt\small \{tanglicheng,liujiaming03,hongzhibin,fanminhu,liujingtuo,dingerrui,wangjingdong\}@baidu.com} \\ 
  {\tt\small frank\_cai@berkeley.edu, mingming.gong@unimelb.edu.au}
\vspace{-5pt}
}
\maketitle

\begin{abstract}
Few-shot font generation (FFG), which aims to generate a new font with a few examples, is gaining increasing attention due to the significant reduction in labor cost.  A typical FFG pipeline considers characters in a standard font library as content glyphs and transfers them to a new target font by extracting style information from the reference glyphs. Most existing solutions explicitly disentangle content and style of reference glyphs globally or component-wisely. However, the style of glyphs mainly lies in the local details, i.e. the styles of radicals, components, and strokes together depict the style of a glyph. Therefore, even a single character can contain different styles distributed over spatial locations. In this paper, we propose a new font generation approach by learning 1) the fine-grained local styles from references, and 2) the spatial correspondence between the content and reference glyphs. Therefore, each spatial location in the content glyph can be assigned with the right fine-grained style. To this end, we adopt cross-attention over the representation of the content glyphs as the queries and the representations of the reference glyphs as the keys and values. 
Instead of explicitly disentangling global or component-wise modeling, the cross-attention mechanism can attend to the right local styles in the reference glyphs and aggregate the reference styles into a fine-grained style representation for the given content glyphs. The experiments show that the proposed method outperforms the state-of-the-art methods in FFG. In particular, the user studies also demonstrate the style consistency of our approach significantly outperforms previous methods.

\end{abstract}

\section{Introduction}
\label{sec:intro}



In the modern era, both computer systems and humans process huge amounts of text information. Fonts, the representations of text, have thus played critical roles in many applications. Therefore, the stylish font generation has its unique commercial and artistic values.
However, building commercial font libraries is costly and labor-intensive. The cost is even higher for those languages with a huge amount of characters (Chinese, Japanese Kanji, Korean, Thai, etc.). 

\begin{figure}   
\centering
\includegraphics[width=\linewidth,scale=1.00]{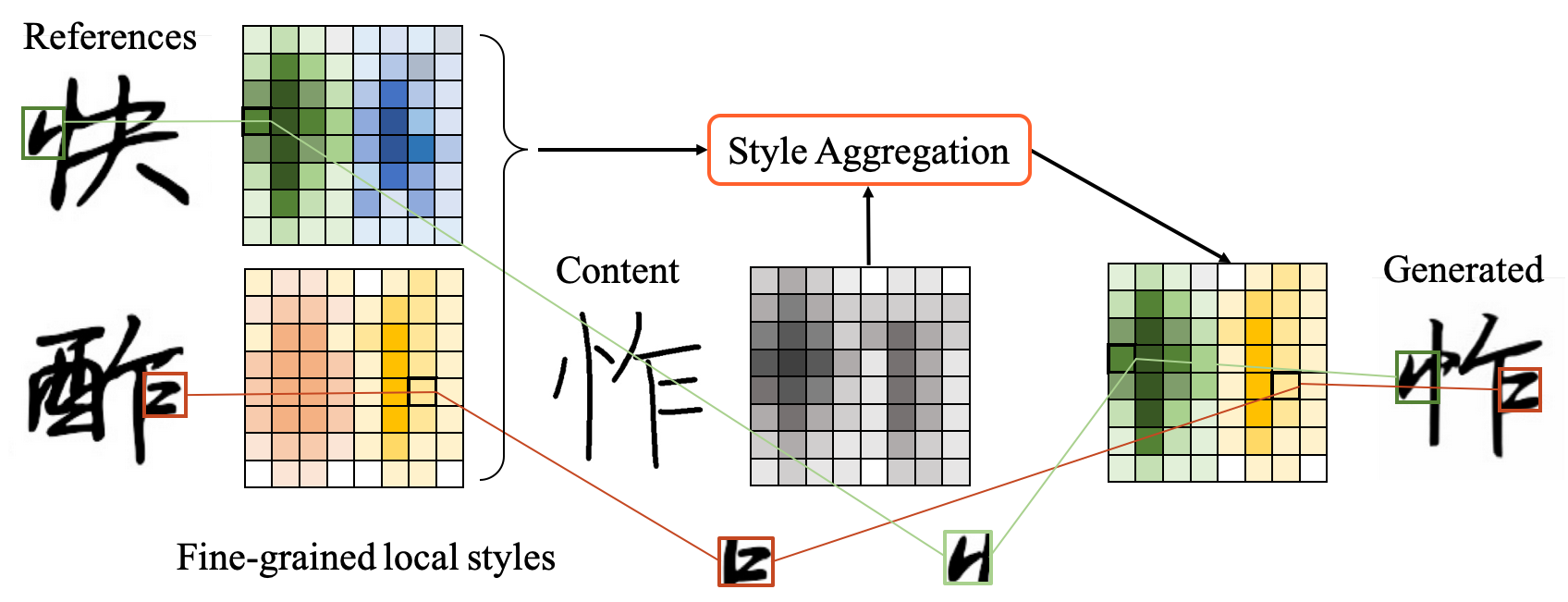}
\caption{Our proposed fine-grained local style extraction and style aggregation process. Our proposed module enables fine-grained style extraction from references and learns the correspondence between content and reference, thus aggregate corresponding local styles into correct locations in content with high-fidelity.}
\vspace{-15pt}


\label{fig1}
\end{figure}

Due to the expert's high cost of building fonts, automatic font generation with deep learning has drawn rising attention. It aims at generating a brand new font library with only a few characters as a reference. With the development of Generative Adversarial Networks (GANs) \cite{goodfellow2014generative, mirza2014conditional}, there have been many classic works of font generation. Early attempts, such as zi2zi \cite{tian2017zi2zi}, use Pix2Pix \cite{isola2017image} like networks with a plug-in font category embedding conditions to learn multiple font styles with a single model. However, these methods require a large number of glyphs to train each unseen font. 


In recent years, some works tried to tackle few-shot font generation (FFG) with a few-shot Image-to-Image translation (I2I) scheme \cite{cha2020few, park2020few, park2021multiple, xie2021dg, gao2019artistic}. Unlike zi2zi, the font style representation is learned from a few reference exemplars, rather than learning embeddings from different font labels. One popular strategy of these works is to explicitly disentangle the content and style representations from given content images and reference exemplars, and two representations are then combined and decoded into the target glyphs. 
With the advance of these works, the generated glyphs' quality is significantly improved when the number of references is limited. Based on the explicit disentanglement ideas, the research of FFG can be divided into two different categories, i.e. global style representation and component-wise style representation. The former one models the glyph style as universal representation for each font \cite{liu2019few, xie2021dg, gao2019artistic}, while the latter one utilizes component-wise style representation from different reference exemplars in the same font \cite{park2021multiple, park2020few, cha2020few, huang2020rd}.

However, in a commercial font, multiple levels of styles need to be considered. An expert would carefully design every possible detail. The detailed styles among components, strokes and, even edges are designed to be consistent. Previous works mostly focus on component-wise styles, while largely ignoring the finer-grained styles. Meanwhile, since the content and style are highly entangled, the commonly used explicit disentanglement can hardly assure the consistency of component-wise styles between the reference glyphs and the generated glyphs. To this end, we employ a references encoder to learn the fine-grained local styles (\textit{FLS}) without explicit disentangled representation learning. Instead of regarding the overall reference map as style, we consider each spatial location of the feature map as a \textit{FLS} representation of reference glyph.  After learning the spatial correspondence between references and content, we further acquire the target style map by aggregating the corresponding \textit{FLS}s from references. Each feature vector of the target style map also represents \textit{FLS}s for the target glyph.

In this paper, we propose a novel approach shown in Figure \ref{fig1}, named FSFont for few-shot font generation. The reference glyph images are encoded into reference maps, which represent the \textit{FLS}s of the references. Our proposed cross-attention based style aggregation module (SAM) learns the spatial correspondence between references and the content glyph. The spatial correspondence is not only on the component level but also on the granular level, which contains more detailed local styles. Afterward, SAM aggregates the \textit{FLS}s of references into the target style map, where each spatial location can be assigned to the right fine-grained style. 
Moreover, to enhance the model to recover details of the references better and learn correspondence more effectively, we adapt a self-reconstruction branch that takes the target character as the input of the reference encoder, and the generated result is supervised by itself. This branch makes learning the correspondence more easily, and helps to produce highly consistent output. Last but not least, we develop a strategy to select the references for each glyph automatically. After analyzing the compositional rules, we design a breadth-first search-based algorithm to search for the reference set and find the optimal references for each character.

In summary, the contribution of this paper is threefold:
\begin{itemize}
\item We devise a novel model for few-shot generation. The model extracts the \textit{FLS} of the reference glyph images, and a cross-attention based style aggregation module aggregates the reference styles into a target style map. The details from reference glyphs are thereby transferred to the target glyph.
\item We propose a unified training framework with a newly designed self-reconstruction branch. This branch significantly boosts the detail capture ability of the model and improves the output images' quality. As a result, the proposed full model achieves state-of-the-art font generation results.
\item We analyze the relationship between characters and select a fairly small set of characters as references. Then we develop a rule to map each character with the elements in the reference set. With the proposed rule, the model's ability to extract component features is better exploited. 
\end{itemize}

\begin{figure*}
\centering
\includegraphics[width=1.03\linewidth,scale=1.00]{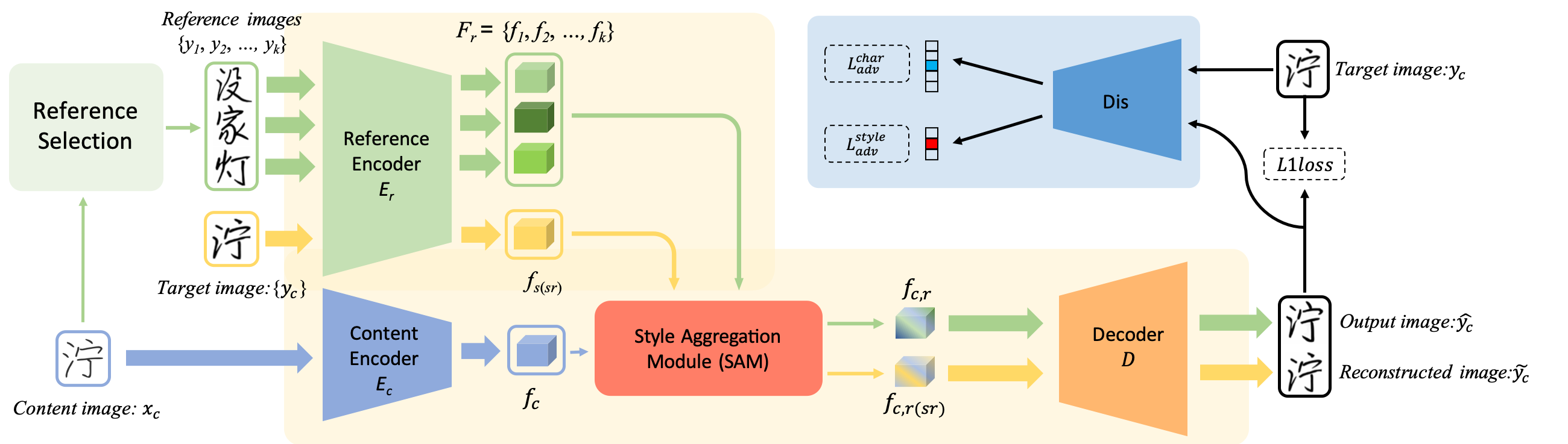}
\caption{\textbf{Overview of our proposed model}. Our generator consists of four parts: a reference encoder $E_r$, a content encoder $E_c$, the \textbf{Style Aggregation Module(SAM)}, and a decoder $D$. Given a content image $x_c$ and $k$-shot references $\{y_1, y_2, ..., y_k\}$ which is selected based on our proposed \textbf{Reference Selection}, $E_r$ and $E_c$ extract their features $F_r$ and $f_c$ respectively. Our SAM matches $F_r$ and $f_c$ based on the attention mechanism and links them with spatial correspondence, outputting the target style map $f_{c,r}$. Afterwards, we use $D$ to obtain the generated image $\hat{y}_c$. We also propose a auxiliary branch of \textbf{Self-Reconstruction} during training stage (yellow branch) . It shares weights with the main branch and improves generated images' quality in details. A multi-task discriminator is employed to calculate the adversarial loss and simultaneously distinguish the content and style category of the generated character. We also compute the pixel-wise reconstruction loss between the ground truth $y_c$ and the generated image $\hat{y}_c$, and between $y_c$ and the reconstructed image $\tilde{y}_c$, respectively.}
\vspace{-5pt}

\label{flowchart}
\end{figure*}

\section{Related Works}
\label{sec:related works}
\subsection{Image-to-image translation}
Image-to-image (I2I) translation refers to the task of learning a mapping function between the source domain and target domain, which preserves the content of the source image while merging the style of the target domain at the same time. According to many I2I methods \cite{isola2017image, zhu2017unpaired, choi2018stargan, choi2020stargan, liu2018unified, yu2019multi, liu2019few}, I2I methods have developed towards multi-mapping \cite{choi2020stargan, yu2019multi, liu2018unified} and few-shot learning \cite{liu2019few}. CycleGAN \cite{zhu2017unpaired} introduces the cycle-consistency into generative models, which enables I2I methods to train cross-domain translation without paired data. FUNIT \cite{liu2019few} accomplish the style transfer task by encoding content and style respectively and combine them with adaptive instance normalization (AdaIN) \cite{huang2017arbitrary}. From an intuitive thought, font generation is a typical I2I translation task, since it tries to keep the content information of the source font and maps it into the target font. Thus, many font generation methods are based on I2I translation methods. 

\subsection{Many-shot font generation}
Early font generation methods \cite{tian2017zi2zi, tian2016rewrite, lyu2017auto, sun2018pyramid, chang2018chinese, jiang2017dcfont, jiang2019scfont, gao2020gan, wu2020calligan, wen2021handwritten, hassan2021unpaired} aim at learning a mapping function between source fonts and target fonts. When new font references are introduced, these methods use hundreds of reference glyphs to fine-tune the original mapping function. zi2zi \cite{tian2017zi2zi} and Rewrite \cite{tian2016rewrite} train GANs in a supervised way with one-hot style labels. AGEN\cite{lyu2017auto} proposes an model based on the auto encoder to transfer standard font images to calligraphy images. HAN \cite{chang2018chinese} designs skip connection and hierarchical loss functions to improve zi2zi's generation performance. These methods require paired data to train a mapping function, e.g,775 for \cite{jiang2019scfont}. Other methods \cite{gao2020gan, hassan2021unpaired} focus on unpaired data, and use them for style extraction. Although these many-shot font generation methods have achieved remarkable performance, it is still a laborious task to collect hundreds of references for the fine-tuning process, especially when the reference font library is glyph-rich.

\subsection{Few-shot font generation}
Most Few-shot font generation (FFG) methods focus on disentangling the content feature and style feature from the given glyphs. Based on different kinds of feature representation, FFG methods can be divided into two main categories: global feature representation \cite{azadi2018multi, zhang2018separating, gao2019artistic, wen2021zigan} and component-based feature representation \cite{huang2020rd, cha2020few, park2020few, park2021multiple}. In methods that apply global feature representation, vectors related to content and style are extracted from content glyphs and reference glyphs respectively. MCGAN \cite{azadi2018multi} synthesize the ornamented glyphs with stacked conditional GANs to extract features from input images. EMD\cite{zhang2018separating} and AGIS-Net\cite{gao2019artistic} combine a style vector and content vector together to synthesize a glyph. ZiGAN\cite{wen2021zigan} matches the features to Hilbert space to better capture the structural information. Works related to component-based feature representation concentrate on devising a feature representation that is related to the glyphs' components or localized features. RD-GAN \cite{huang2020rd} uses a radical encoder to extract features of glyphs' specific components. In DM-Font \cite{cha2020few}, it disassembles glyphs to stylized components and reassembles them to new glyphs by utilizing strong compositionality prior. LF-Font \cite{park2020few} designs a component-conditioned reference encoder to extract component-wise features from reference images. MX-Font \cite{park2021multiple} employs multiple encoders for each reference image with disentanglement between content and style which makes the cross-lingual task possible. DG-Font \cite{xie2021dg} is an unsupervised framework based on TUNIT \cite{Baek_2021_ICCV} by replacing the traditional convolutional blocks with Deformable blocks which enables the model to perform better on cursive characters which are more difficult to generate.


However, the previous FFG works fail to fully explore the style maps from $k$-shot references. When receiving $k$-shot reference images, they tend to explicitly disentangle $style$ and $content$ of images globally or component-wisely and conduct a average operation among extracted features \cite{park2020few, park2021multiple}. 
The local details extracted from each reference are significantly weakened by disentanglement and the average operation.
Therefore, we design \textit{Style Aggregation Module} that aims to keep the very detailed features from reference images and to fully utilize the spatial details.

\subsection{Attention Mechanism}
The attention mechanism \cite{vaswani2017attention} is known to capture dependence. After its debut in machine translation, it has been applied in many vision tasks including font generation. RD-GAN\cite{huang2020rd} using attention mechanism to extract rough radicals from content characters and then render them into target style. HWT \cite{bhunia2021handwriting} uses transformer blocks to bridge the gap between image and text, making it capable of generating stylized handwriting English text images. Our
\textit{Style Aggregation Module} is highly motivated by the attention mechanism for fine-grained feature map re-composition.
\section{Method}
In this section, we present the details of our FS-Font method. We first briefly review the few-shot font generation problem setup and introduce the overall framework of our method (\ref{Sec:3.0}). Next, we present the details of three crucial components of our approach, including the \textit{Style Aggregation Module} (SAM) (Sec. \ref{sec:3.1}), the Self-Reconstruction branch (Sec. \ref{sec:3.2}), and Reference Selection (Sec. \ref{sec:3.3}). 

\subsection{Problem Setting and Method Overview}\label{Sec:3.0}
In few-shot font generation, given a content glyph image $x_c$ with the content $c$ from a standard font $X = \{x_c\}^N_{c=1}$ and a $k$-shot reference glyph images with the style $s$: $R_{s}=\left\{y_i\right\}_{i=1}^k\subset Y_s$, our goal is to generate a stylized image $\hat{y}_c$ that has the content $c$ and style $s$ via a generator $G$:
\begin{equation}
\label{eq:1}
\hat{y}_c=G(x_c, R_{s}),
\end{equation} 
where the style $s$ of $\hat{y}_c$ is omitted for sake of simplicity. 

In training, we collect $L$ fonts with different styles $Y = \{Y_{s}\}^L_{s=1}$, where $Y_{s} = \{y_c\}^N_{c=1}$. Therefore, we have a paired dataset $\{x_c,y_c\}_{c=1}^N$ for each content $c$ in $s$-th training style.

The overall framework is shown in Figure \ref{flowchart}. The reference encoder $E_r$ first encodes $R_s$ into $k$-shot reference maps $F_r = \left\{f_{i}\right\}_{i=1}^k$, where $f_i$ is encoded from $y_i$. The content encoder $E_c$ extracts the content feature map $f_c$ from the input content image $x_c$. Our proposed SAM takes $F_r$ and $f_c$ as inputs, attends to the corresponding spatially local styles in the reference maps $F_r$ and aggregates the local styles into the target style map $f_{c, r} = SAM(f_c, F_r)$. In the end, the decoder $D$ decodes $f_{c, r}$ into the generated output image $\hat{y}_{c}$. A multi-task projection discriminator\cite{miyato2018cgans} is employed to discriminate each generated image and real image. The discriminator outputs a binary classification of fake or real for each character's style and content category.



\subsection{Style Aggregation Module}
\label{sec:3.1}

\begin{figure}   
\centering
\includegraphics[width=\linewidth,scale=1.00]{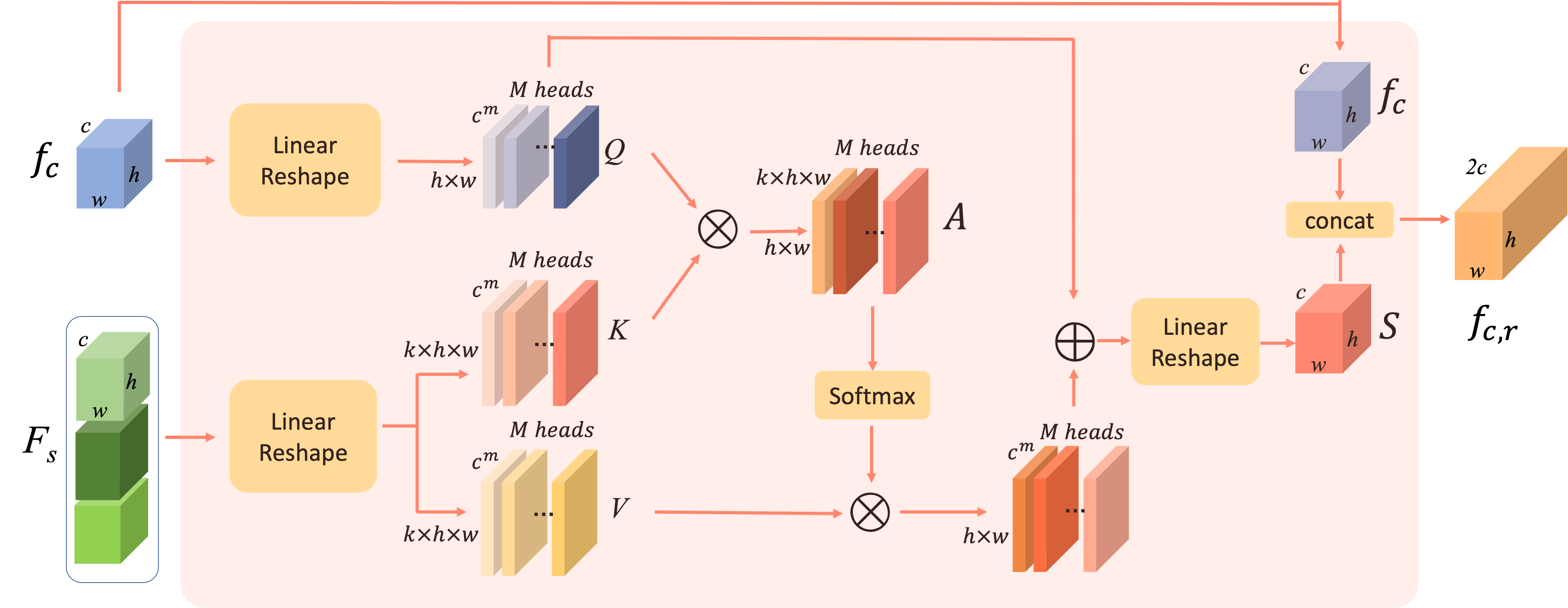}
\caption{\textbf{The illustration of the Style Aggregation Module (SAM)}. We use multihead attention mechanism to calculate the spatial correspondence of content and references and generate the fine-grained feature map.}
 \vspace{-0.1in}

\label{fig:attention}
\end{figure}

An overview of the SAM is depicted in Figure \ref{fig:attention}. The core in SAM is a multi-head cross attention block that attends to spatially local styles from the reference maps $F_r$ and aggregates the reference styles into the fine-grained style representation for the given content image. For the $m$-th attention head, SAM learns a Query map $Q^m$ from content feature map $f_c$ and Key map $K^m$ from reference maps $F_r$, which achieves the spatial correspondence matrix $A^m$ between the pixels of $Q^m$ and $K^m$. A value map $V^m$ is simultaneously learned from $F_r$. Multiplying the correspondence matrix $A^m$ with the value map $V^m$ aggregates the local style styles into the target style $S^m$. As different head captures different information, we combine all the target styles together for decoding.

Formally, we reshape $f_c \in \mathbb{R}^ {c \times h \times w}$ into a sequence $\tilde{f_c} \in \mathbb{R}^ {c \times (h \cdot w)}$, where $(h, w)$ is the resolution of the feature map, $c$ is the number of channels. For the $m$-th head, after applying a linear layer $L^m_{query} \in \mathbb{R}^{c \times c^m}$, we acquire the query matrix $Q^m \in \mathbb{R}^{c^m\times (h \cdot w)}$. Meanwhile, the reference maps $F_r = \left\{f_{i}\right\}_{i=1}^k$ are reshaped and concatenated along the spatial dimension $(h, w)$, forming a reference sequence $\tilde{f}_r \in \mathbb{R}^ {c \times (k \cdot h \cdot w)}$. We multiply $\tilde{f}_r$ by two linear projections $L^m_{key}, L^m_{value} \in \mathbb{R}^{c \times c^m}$ and generate a key map $K^m$ and a value map $V^m$ as follows: 
\begin{equation}
\begin{split}
  &Q^m = L^m_{query}(\tilde{f_c}), \quad Q^m \in \mathbb{R}^{c^m \times (h \cdot w)}, \\
  &K^m = L^m_{key}(\tilde{f_r}),   \quad K^m \in \mathbb{R}^{c^m \times (k \cdot h \cdot w)}, \\
  &V^m = L^m_{value}(\tilde{f_r}), \quad V^m \in \mathbb{R}^{c^m \times (k \cdot h \cdot w)}. \\
  \end{split}
  \label{eq:key matrix}
\end{equation} 

We then compute a spatial correspondence matrix $A^m$ of which each element $A^m(u, v)$ is a pairwise feature correlation between content feature in position $u$ and reference feature in position $v$, calculated as follows: 

\begin{equation}
  A^m = \frac{Q^{m \top} K^m}{\sqrt{c^m}} \in \mathbb{R}^{hw \times khw},
  \label{eq:key matrix}
\end{equation}
where $c^m$ is the hidden dimension of $Q^m$ and $K^m$. The $1/\sqrt{c^m}$ factor follows Transformers\cite{vaswani2017attention} to prevent the magnitude of the dot product from growing extreme.

With the correspondence matrix $A^m$, we obtain the aggregated style from references by 
\begin{equation}
  S^m = softmax(A^m) V^{m\top} \in \mathbb{R}^{hw \times c^m}.
  \label{eq:key matrix}
\end{equation}

After permuting and reshaping $S^m$ into $\mathbb{R} ^ {c^m \times h \times w}$, we concatenate all $S^m$ along the channel dimension, and employ a linear projection $L_s \in \mathbb{R}^{(c^m \cdot M) \times c}$ to obtain $S$. 
The target style map $f_{c,r}$ is obtained as the concatenation between $S$ and content feature $f_c$. The decoder $D$ decodes it into the target image $\hat{y}_{c}$ as follows:

\begin{equation}
\begin{split}
  &S = L_s(S^1 \circ S^2, ..., \circ S^M) \in \mathbb{R}^{c \times h \times w},\\  
  &\hat{y}_c = D(f_{c,r}) = D(f_c \circ S),
  \label{eq:decode}
\end{split}
\end{equation}
where $\circ$ denotes concatenation operator and $M$ is the number of total attention heads.

\subsection{Self Reconstruction}
\label{sec:3.2}
To achieve a highly style-consistent glyph using SAM, it depends not only on the proper aggregation of styles, but also on the expressivity of the styles that depict fine and local details of the references. However,
the $k$-shot font generation setup is a hard-to-learn problem, as it requires the attention learns spatial correspondence from weakly correlated content-references pairs, which is sometimes confusing even for a human expert. 
Thus, besides the main branch of $k$-shot learning, we introduce an easy-to-learn Self-Reconstruction (SR) branch to boost the learning process. 

While the generator is trained in the above mentioned $k$-shot setup, for a given content input $x_c$, we have the ground truth image $y_c$ to supervise the model training. Self Reconstruction is a parallel branch that shares the same model as the main branch during training. It takes $y_c$ as the reference image $\tilde{R_s}=\left\{y_c\right\}$, and its output $\tilde{y}_{c}$ is also supervised by $y_c$ itself. In contrast to Eq. \ref{eq:1}, the generation process of this branch is 
\begin{equation}
\tilde{y}_{c}=G(x_c, \tilde{R_s}).
\end{equation}

The detailed training setup can be found in Sec.\ref{sec:3.4}. In the SR branch, the content and reference images are strongly correlated. The spatial correspondence matrix can be easily learned as the strokes and the components' relationship between content and reference is clear. With the well-learned correspondence, the generator can be optimized with well-aligned gradients. As a result, the expressivity of depicting details can be further learned within our framework.

\subsection{Reference Selection}
\label{sec:3.3}

\begin{figure}   
\centering
\includegraphics[width=\linewidth,scale=1.0]{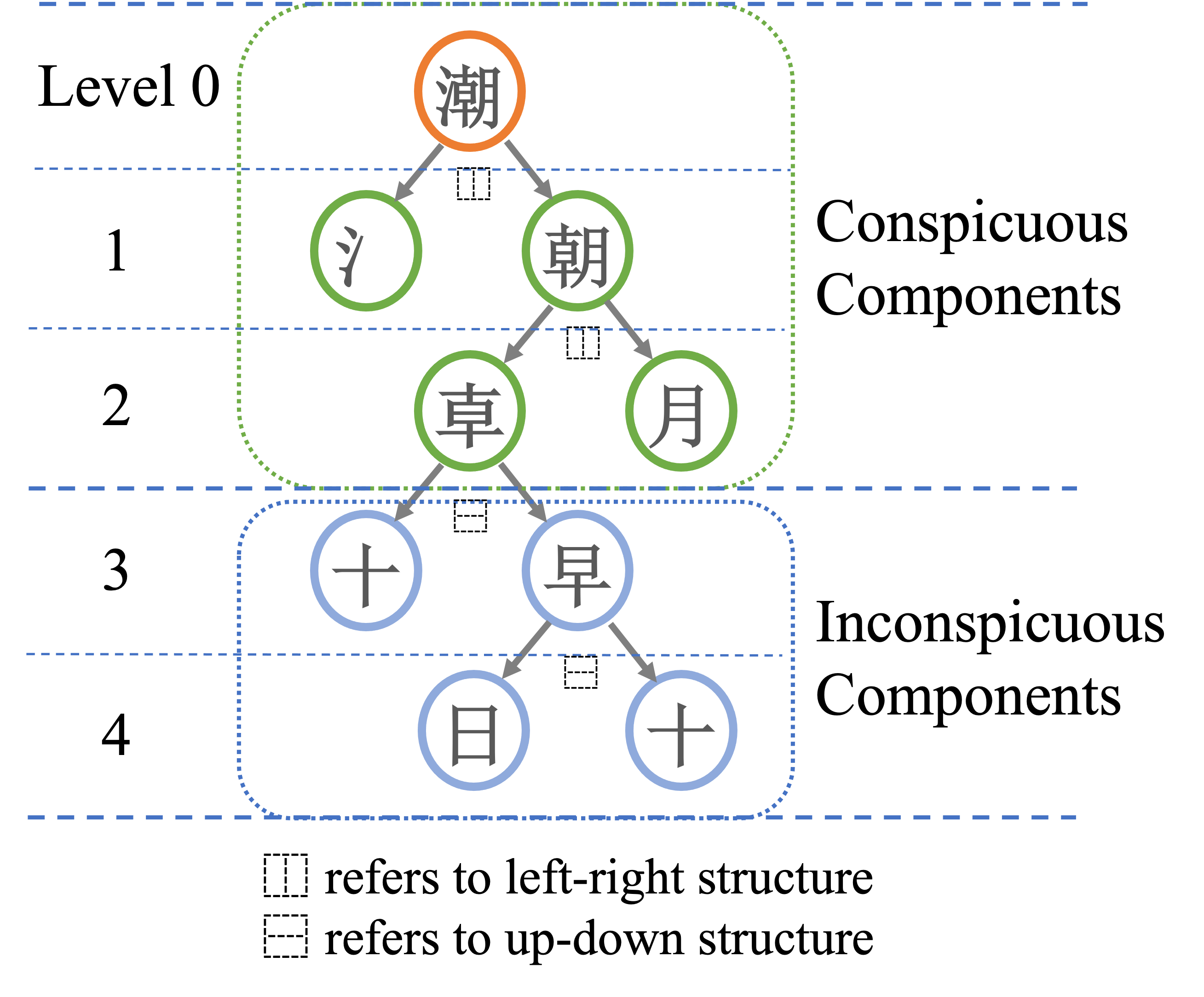}
\caption{A component tree. 
To make the mapping process more sensible, the component's structure information is also taken into consideration.}
\vspace{-10pt}
\label{tree_analyzer}
\end{figure}

In previous works considering the decomposition of components like LF-Font\cite{park2020few}, reference characters that contain the same components as content character are randomly selected from the training set during each iteration. The model can hardly learn how to extract the right component-wise features with varying combinations of the reference set. Thus, we introduce a strategy to select a fixed reference set whose components cover most of the commonly used characters and design a content-reference mapping that fixes the combination of reference set for each character. To establish this mapping function, we firstly decompose each character into a component tree, as shown in Figure \ref{tree_analyzer}, based on a commonly used decomposition table\footnote{https://github.com/cjkvi/cjkvi-ids/blob/master/ids.txt}.  We define the components at the level 0, 1, and 2 as the conspicuous components, which contains both radical and compositional structures that easier be transfered from the references to the target. 

\textbf{Reference set selection}. This reference set should cover as many conspicuous-level components as possible. Initially, we select a small subset (typically including 100 characters). First, we decompose the characters into component trees. Once the character contains two or more new components, we add this character to our reference set. When the elements in the reference set reach their limits, we stop searching and obtain the style reference set and its corresponding contained components.

\textbf{Content-reference mapping}. After completing the reference set selection, we then establish the mapping relations between content glyphs and style references. We propose a greedy process to find $k$-shot references for a glyph. In this process, we search the reference set for $k$ times to establish a mapping relation. During every searching step, we find the reference glyph that shares the most components with the target glyph. If there are multiple solutions, we select the optimal solution that has the most components with the same structure composition. After selection, we remove the reference from the reference set and continue the next searching step. By this process, we can determine every glyph's corresponding $k$-shot references.

\subsection{Training}
\label{sec:3.4}
We train our model to generate the image $\hat{y}_{c}$ from a content glyph image $x_c$ and a fixed set of reference glyph images $R_s$. In each iteration, the main branch and the SR branch generate $\hat{y}_{c}$ and $\tilde{y}_{c}$ simultaneously and are supervised by the same losses. \textbf{FSFont} learns the reference encoder $E_r$, content encoder $E_c$, Style Aggregation Module and decoder $D$ with following losses: 1) Adversarial loss with the multi-task discriminator. 2) L1 loss among $\hat{y}_{c}$, $\tilde{y}_{c}$ and a paired ground truth image ${y_c}$. 

\textbf{Adversarial loss}. Since we our aim is to generate visually high-quality images, we employ a multi-head projection discriminator \cite{park2020few} in our framework. The loss function is represented as follows:

\begin{equation}
\begin{split}
  \mathcal{L}_{adv}^D & = \mathbb{E}_{y_c \in P_{d}}[min(0,1-D_{s, c}(y_c))] \\
  &+ \mathbb{E}_{\bar{y}_c\in P_{g}}[min(0,-1-D_{s, c}(\bar{y}_{c}))] ,\\
  \mathcal{L}_{adv}^G & = -\mathbb{E}_{\bar{y}_c\in P_{g}}D_{s,c}(\bar{y}_{c}),
  \label{eq GAN loss}
\end{split}
\end{equation}
where $\bar{y}_{c} \in \{\hat{y}_{c}, \tilde{y}_{c}\}$, $D_{s,c}(\cdot)$ refers to the logits from the character $c$ head and style $s$ head from the discriminator, $P_g$ denotes the set of generated images from both main branch and Self-Reconstruction branch, and $P_d$ denotes the set of real glyph images. \\

\textbf{L1 loss}. For learning the pixel-level consistency, we employ an L1 loss between the generated images $\hat{y}_{c}$, $\tilde{y}_c$ and ground truth image $y_c$. 
\begin{equation}
  \mathcal{L}_{l1} = \mathbb{E}\left[\lVert \hat{y}_{c} - y_c\rVert_{1} + \lVert \tilde{y}_{c} - y_c\rVert_{1}\right].
  \label{eq l1 loss}
\end{equation}

\textbf{Overall objective loss}.
Combining all losses mentioned above, we train the whole model by the following objective:
\begin{equation}
\begin{split}
   \mathop{min}\limits_{G}\mathop{max}\limits_{D}\lambda_{adv}(\mathcal{L}_{adv}^G + \mathcal{L}_{adv}^D) + \lambda_{l1}\mathcal{L}_{l1}, 
\label{eq:key matrix}
\end{split}
\end{equation}
where $\lambda_{adv}$ and $\lambda_{l1}$ are hyperparameters to control the weight of each loss. We empirically set  $\lambda_{adv}=1$ and $\lambda_{l1}=0.1$ in our experiments.

\section{Experiments}
In this section, we evaluate FSFont for the representative Chinese font generation task. We first introduce the datasets we used and compare our framework with other state-of-the-art(SOTA) methods. After that,  an ablation study evaluates the effectiveness of each module in our framework and how they affect final results.

\begin{table*}
  \setlength\tabcolsep{15pt}
  \centering
  \begin{tabular}{l|cccccc}
      \toprule 
      \multicolumn{7}{c}{\textbf{Unseen Fonts Seen Characters}} \\ \cmidrule(lr){1-7}
      Methods  & L1 loss $\downarrow$ & RMSE $\downarrow$ & SSIM $\uparrow$ & LPIPS $\downarrow$ & User(C) \% $\uparrow$ & User(S) \% $\uparrow$ \\
      \midrule
      FUNIT \cite{liu2019few}  & 0.148 & 0.344 & 0.565 & 0.2543 & 86.0 & 0.3\\
      DG-FONT \cite{xie2021dg}  & 0.131 & 0.329 & 0.604 & 0.2154 & 92.4 & 6.1\\
      MX-FONT \cite{park2021multiple}  & 0.152 & 0.347 & 0.584 & 0.2291 & \textbf{97.4} & 8.4 \\
      AGIS-NET \cite{gao2019artistic}  & 0.105 & 0.289 & 0.651 & 0.1865 & 97.2 & 5.2\\
      LF-FONT \cite{park2020few}  & 0.129 & 0.322 & 0.607 & 0.2006 & 93.4 & 11.3\\
      Ours  & \textbf{0.097} & \textbf{0.268} & \textbf{0.671}& \textbf{0.1618} & 96.6 & \textbf{68.7}\\
      
      \midrule
      \multicolumn{7}{c}{\textbf{Unseen Fonts Unseen Characters}} \\ \cmidrule(lr){1-7}
       Methods  & L1 loss $\downarrow$ & RMSE $\downarrow$ & SSIM $\uparrow$ & LPIPS $\downarrow$ & User(C) \% $\uparrow$ & User(S) \% $\uparrow$ \\ 
      \midrule
      FUNIT \cite{liu2019few}  & 0.152 & 0.345 & 0.532 & 0.2424 & 84.6 & 1.2 \\
      DG-FONT \cite{xie2021dg}  & 0.141 & 0.341 & 0.573 & 0.2151 & 86.4 & 3.1\\
      MX-FONT \cite{park2021multiple}  & 0.153 & 0.352 & 0.573 & 0.2317 & 93.2 & 11.2\\
      AGIS-NET \cite{gao2019artistic}  & 0.114 & 0.302 & 0.623 & 0.1877 & 89.8 & 4.1\\
      LF-FONT \cite{park2020few}  & 0.138 & 0.334 & 0.577 & 0.2018 & 90.6 & 13.3\\
      Ours  & \textbf{0.106} & \textbf{0.283} & \textbf{0.642} & \textbf{0.1627} & \textbf{93.8} & \textbf{67.1}\\
      
      \bottomrule 
  \end{tabular}
  \vspace{-0.05in}
  \caption{\textbf{Qualitative comparison on UFUC and UFSC datasets}}
  \label{comparison table}
\end{table*}

\begin{figure*}  
\centering
\includegraphics[width=\linewidth,scale=1.00]{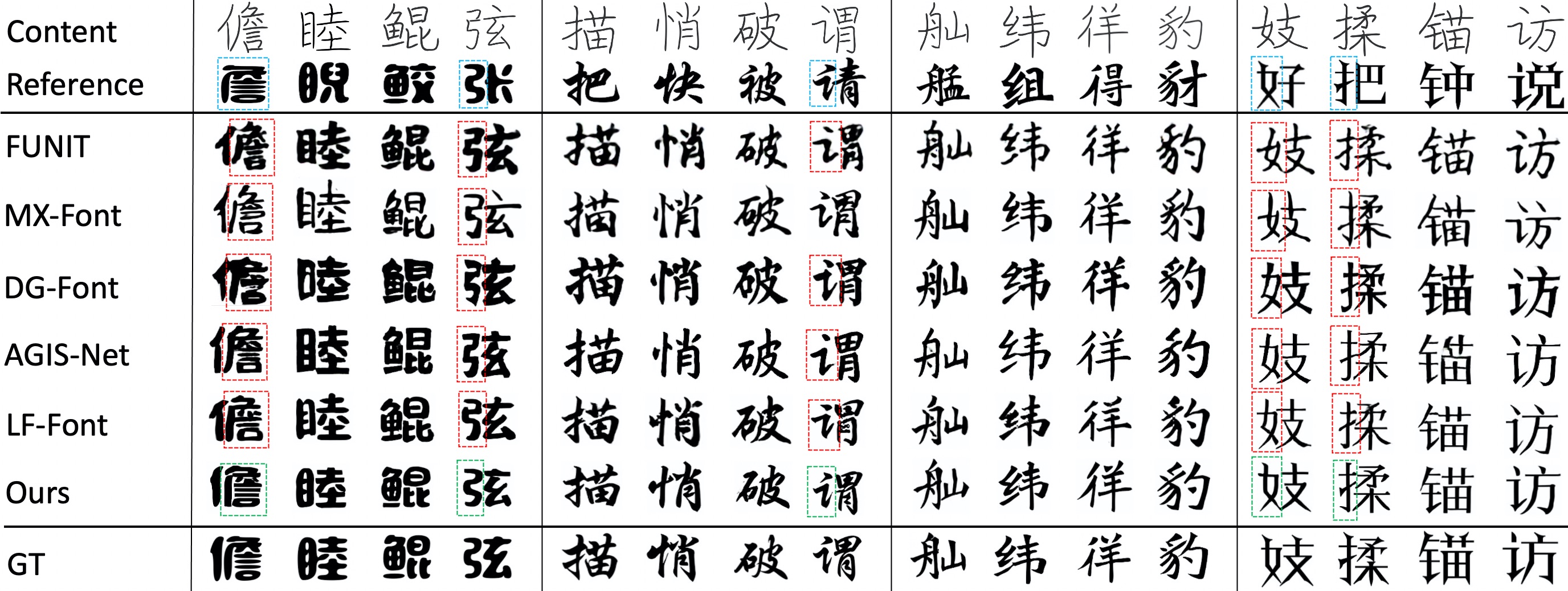}
\caption{\textbf{Generated results of each method on UFUC datasets.} We represent the generated samples of four different kinds of fonts. We mark the main component of each character with boxes. The blue boxes in Reference are components we hope the model recover. It shows our model(green boxes) is capable of recovering more details from Reference than other SOTA works do(red boxes).}
\vspace{-0.1in}

\label{qualitative comparison}
\end{figure*}

\subsection{Datasets and evaluation metrics}
\label{sec:4.1}
\textbf{Datasets}. We choose 407 fonts and 3396 commonly used Chinese characters as our datasets including handwriting fonts, printed fonts as well as artistic fonts. All images are $128\times128$ pixels. We select 100 characters from datasets as our reference set and create a Content-Reference mapping with the strategy discussed in Sec.\ref{sec:3.3}. Reference set and Content-Reference mapping are fixed in both training set and testing set which means only 100 characters are needed to generate a new font library. The training set contains 397 fonts, 2896 characters. The test set are from 10 representative fonts including typewriter fonts, artistic fonts, and handwriting fonts to evaluate the generalization of our model on variant unseen fonts. We test the methods with two setups \textit{Unseen Fonts Unseen Characters}(UFUC) and \textit{Unseen Fonts Seen Characters}(UFSC). UFUC refers to the 500 characters in the test fonts, and UFSC refers to the 2896 characters in the test fonts.

\textbf{Evaluation metrics}. 
To evaluate the similarity between generated images and ground truth, we compare our framework with other SOTA methods in the following metrics, i.e. \textbf{L1}, \textbf{RMSE}, \textbf{SSIM} and \textbf{LPIPS}. Additionally, we conduct user studies to calculate the \textbf{Character Accuracy}, as CNN-based classifiers are tolerant of little defects like missing or broken strokes and blurry edges in a stroke-rich character. We hire 51 volunteers to rigorously count the correct ones from 500 generated characters of each method. A character will be counted as correct only if the volunteer can not spot a defect. For \textbf{Style Consistency}, the volunteers are required to evaluate which of the methods generates the most similar character given a set of reference glyph images from 30 randomly selected cases. 

\subsection{Comparison methods}
We compare our method with previous SOTA methods. 1) \textbf{FUNIT}\cite{liu2019few} is a early work on Few-shot image translation in an Unsupervised way which introduces two different encoders and an AdaIN module to generate a new image with mixed content and style. 
2) \textbf{DG-Font}\cite{xie2021dg} is an Unsupervised network using Deformable Convolution in Generator to achieve a better effect on cursive characters  3) \textbf{MX-Font}\cite{park2021multiple} adopts multiple experts to extract different local structures which makes cross-lingual generation task possible. (4) \textbf{AGIS-net}\cite{gao2019artistic} uses two different decoders to generate images with shape and texture information which makes generated image more stable. (5) \textbf{LF-Font} \cite{park2020few} proposes localized style representation which makes it enable to extract the component-wise features. For a fair comparison, we choose the Kaiti Line Font $\footnote{
Font is available at https://chanind.github.io/hanzi-writer-data/}$ as the standard font and re-train all models on the training datasets in Sec. \ref{sec:4.1}. 

\subsection{Experimental results}
\textbf{Quantitative comparison}. Table \ref{comparison table} shows the FFG performance of our \textbf{FSFont} and other competitors. We conducted the experiment on UFUC and UFSC. FSFont clearly outweighs previous SOTA methods on all of the similarity metrics from pixel-level to perceptual-level. On UFSC setup, MX-Font\cite{park2021multiple} and AGIS-Net \cite{gao2019artistic} generates little more correct characters than FSFont. However, FSFont still generates the most correct characters on UFUC. 
AGIS-Net suffers from a performance drop on \textbf{Character Accuracy} in UFUC. This may suggest that its generalization to new characters are limited. Meanwhile, the user study of \textbf{Style Consistency} of FSFont is remarkably better than other methods, which further verifies that FSFont generates the visually similar results from users' perspective.

\textbf{Qualitative comparison}
We illustrate the generated samples in Figure \ref{qualitative comparison} for each method. We selected four different fonts including typewriter fonts, artistic fonts as well as handwriting fonts to see the generalization of all competitors. As demonstrated in Figure \ref{qualitative comparison}, FSFont are available to recover as many details from reference images. Though other methods like LF-Font \cite{park2020few} or AGIS-Net \cite{gao2019artistic} are able to generate stable characters and recover coarse features such as the thickness of strokes and inclination of font, they could not recover details as explicitly as our method does.

\begin{table}
  \setlength\tabcolsep{5pt}
  \begin{tabular}{cccccccc}
      \toprule 
      \multicolumn{7}{c}{\textbf{Unseen Fonts Unseen Characters}} \\ \cmidrule(lr){1-7}
      SAM & SR & RS & L1 loss $\downarrow$ & RMSE $\downarrow$ & SSIM $\uparrow$ & LPIPS $\downarrow$ \\
      \midrule
      \color{green}\Checkmark & \color{green}\Checkmark & \color{green}\Checkmark & \textbf{0.106} & \textbf{0.283} & \textbf{0.642} & \textbf{0.1627} \\
      \color{red}\XSolidBrush & \color{green}\Checkmark & \color{green}\Checkmark & 0.113 & 0.294 & 0.621 & 0.1822 \\
      \color{green}\Checkmark & \color{red}\XSolidBrush & \color{green}\Checkmark & 0.127 & 0.318 & 0.596 & 0.1859 \\
      \color{green}\Checkmark & \color{green}\Checkmark & \color{red}\XSolidBrush & 0.114 & 0.292 & 0.624 & 0.1798 \\
      \color{red}\XSolidBrush & \color{red}\XSolidBrush & \color{red}\XSolidBrush & 0.131 & 0.338 & 0.584 & 0.2189
      \\
      
      \bottomrule 
      \end{tabular}
  \caption{\textbf{Analysis of different modules in our proposed framework.} By discarding the Style Aggregation Module (SAM), Self Construction (SR), and Reference Selection (RS) individually, we can see that all these modules have positive effects on the original model respectively. The model with three modules has the best performance in all evaluation metrics.}
  \vspace{-0.2in}
  \label{overall ablation}
\end{table}

\begin{table}
  \setlength\tabcolsep{4pt}
  \begin{tabular}{lcccccc}
      \multicolumn{6}{c}{} \\
      \raisebox{+.9\height}{Content} & \includegraphics[width=0.1\linewidth,scale=1.00]{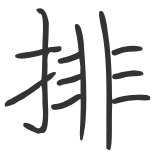} &  \includegraphics[width=0.1\linewidth,scale=1.00]{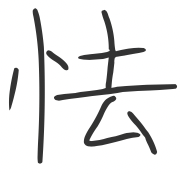} &  \includegraphics[width=0.1\linewidth,scale=1.00]{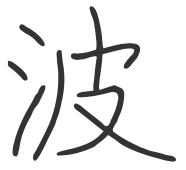} &  \includegraphics[width=0.1\linewidth,scale=1.00]{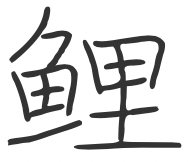} &  \includegraphics[width=0.1\linewidth,scale=1.00]{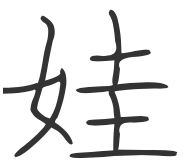} &  \includegraphics[width=0.1\linewidth,scale=1.00]{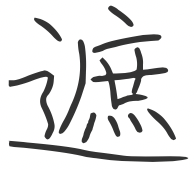} \\
      
      \raisebox{+.7\height}{Reference} & \includegraphics[width=0.1\linewidth,scale=1.00]{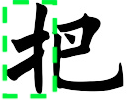} &  \includegraphics[width=0.1\linewidth,scale=1.00]{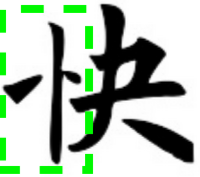} &  \includegraphics[width=0.1\linewidth,scale=1.00]{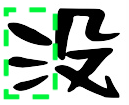} &  \includegraphics[width=0.1\linewidth,scale=1.00]{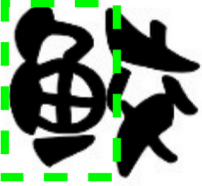} &  \includegraphics[width=0.1\linewidth,scale=1.00]{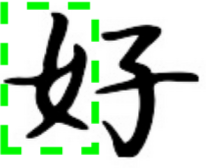} &  \includegraphics[width=0.1\linewidth,scale=1.00]{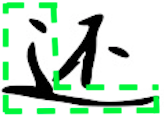} \\
      
      \midrule
       \raisebox{+.7\height}{w/o SR} & \raisebox{-.0\height}{\includegraphics[width=0.1\linewidth,scale=1.00]{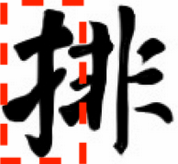}} &  \raisebox{-.0\height}{\includegraphics[width=0.1\linewidth,scale=1.00]{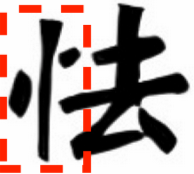}} &  \includegraphics[width=0.1\linewidth,scale=1.00]{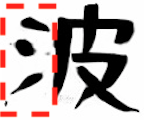} &  \includegraphics[width=0.1\linewidth,scale=1.00]{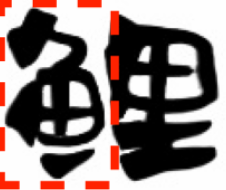} &  \includegraphics[width=0.1\linewidth,scale=1.00]{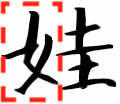} &  \includegraphics[width=0.1\linewidth,scale=1.00]{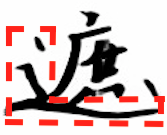}  \\

      \raisebox{+.7\height}{with SR} & \includegraphics[width=0.11\linewidth,scale=1.00]{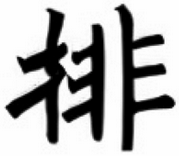} &  \raisebox{-.0\height}{\includegraphics[width=0.1\linewidth,scale=1.00]{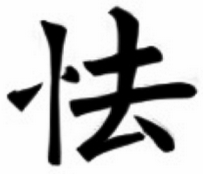}} &  \includegraphics[width=0.1\linewidth,scale=1.00]{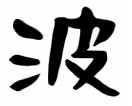} &  \includegraphics[width=0.1\linewidth,scale=1.00]{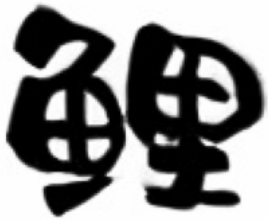} &  \includegraphics[width=0.1\linewidth,scale=1.00]{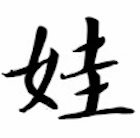} &  \includegraphics[width=0.1\linewidth,scale=1.00]{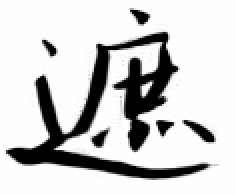} \\

      \raisebox{+.7\height}{GT} & \raisebox{-.1\height}{\includegraphics[width=0.11\linewidth,scale=1.00]{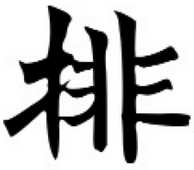}} &   \raisebox{-.0\height}{\includegraphics[width=0.1\linewidth,scale=1.00]{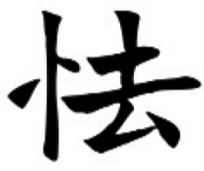}} &  \raisebox{-.1\height}{\includegraphics[width=0.1\linewidth,scale=1.00]{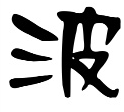}} &  \includegraphics[width=0.1\linewidth,scale=1.00]{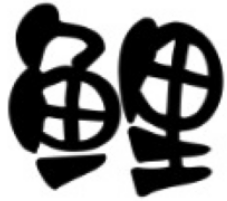} &  \includegraphics[width=0.1\linewidth,scale=1.00]{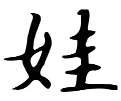} &  \raisebox{-.1\height}{\includegraphics[width=0.1\linewidth,scale=1.00]{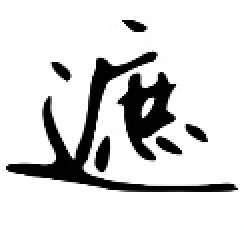}} \\
      
  \end{tabular}
\caption{\textbf{Comparison of generated images with and without Self-Reconstruction branch}. The green box in the reference set is the component we hope the model to recover. The red box shows the insufficient details generated from the model trained without Self Reconstruction branch.}
\vspace{-0.1in}
\label{SR ablation}
\end{table}

\begin{figure*}   
\centering
\includegraphics[width=\linewidth,scale=1.0]{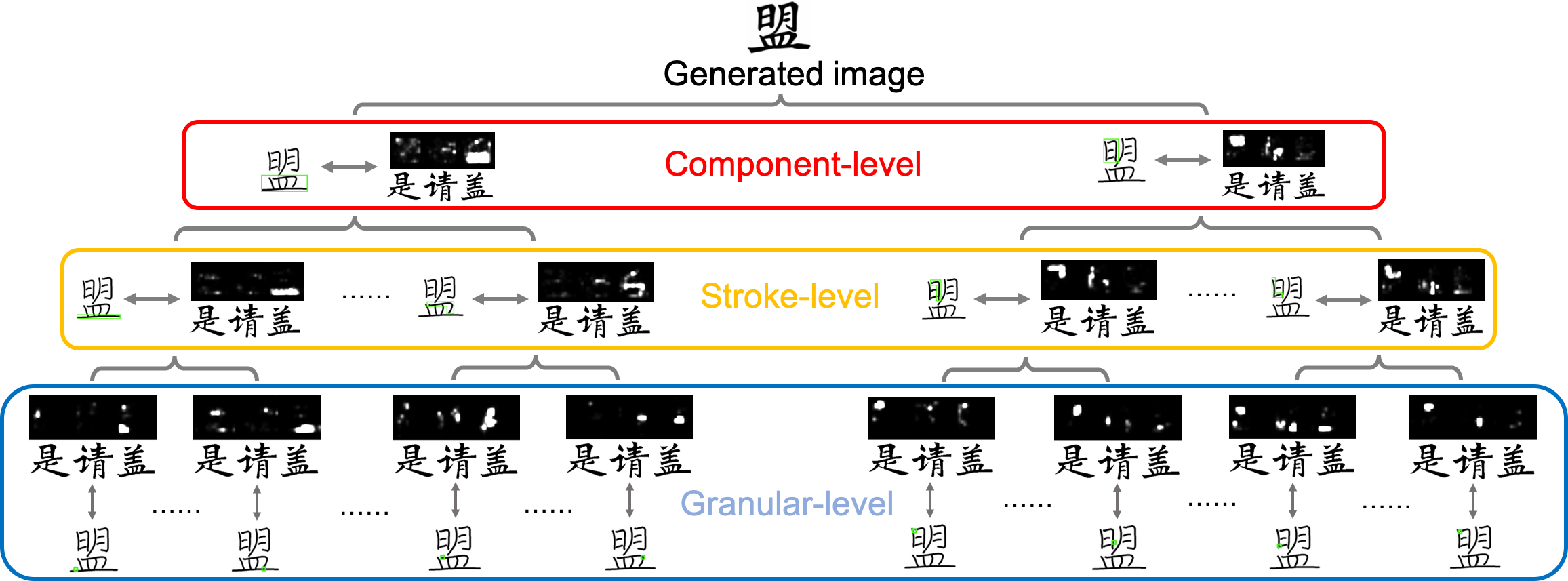}
\caption{\textbf{Visualization of Style Aggregation Module.} The brighter spot in the attention maps denotes the larger contribution of corresponding features in the reference feature maps. The light-green dot, line and bounding box denote the queries from different level. 
}
\vspace{-0.1in}
\label{FRB ablation}
\end{figure*}

\textbf{Ablation studies}.
In this part, we discuss the effectiveness of each module we introduce in \textbf{FSFont}. We discard each module at a time and train the model with other settings unchanged. The overall evaluation results on UFUC datasets are shown in Table \ref{overall ablation}. We replace SAM with averaging features in $F_r$ to test its effect. For Reference Selection and Content-Reference mapping, we replace them with the strategy of LF-Font\cite{park2020few} by randomly selecting reference glyph images with a common component set for each content character. Both two modules have a positive effect on the final results. The Self-Reconstruction Branch has a significant impact on outputs. As shown in Table \ref{SR ablation}, the model trained without SR branch can hardly recover details from reference glyph images. 

\textbf{Visualization of SAM}. To demonstrate the effectiveness of SAM, we visualize the attention maps for different levels in Figure \ref{FRB ablation}. 
Specifically, given a certain spatial point $q$ in the content feature map as a query, we can obtain the corresponding correlations $A^m_q \in \mathbb{R}^{khw}$ from the spatial correspondence matrix $A^m \in \mathbb{R}^{hw \times khw}$, and construct an attention map by reshaping $A^m_q$ to $h \times kw$. We consider the queries from the \textbf{Granular}, \textbf{Stroke}, and \textbf{Component-level}, respectively, and compute the final attention map by summing over the attention maps related to the queries.  
It shows that our SAM module empowers the model to attend to the correct \textit{FLS}s from reference images and extract a sub-component level feature representation for content images.

\section{Conclusion}
In this paper, we propose a novel FFG model, which is able to calculate the spatial correspondence of the content and reference based on their component features. Our proposed Style Aggregation Module aggregates fine-grained local styles of references to content's corresponding location with high-fidelity. Besides, we propose a Self-Reconstruction branch to help model to recover details from references. Last but not least, our Reference Selection strategy guarantee that each content can match references that share common conspicuous components. Our extensive experiments show that FSFont significantly outperforms other methods in both objective and subjective similarity. \textbf{Limitation} The model is trained on limited data, it can not faithfully replicate every detail of the font. \textbf{Negative Impact} Though FSFont can be potentially used to imitate handwriting, a human expert can still spot the difference between generated and real handwritings. 
{\small

\bibliographystyle{ieee_fullname}
\bibliography{ms}
}

\clearpage

\appendix

{\section*{Appendices}\huge }
\section{Implementation Details}

\subsection{Reference Selection}
In this section, we will discuss the implementation details of reference selection, which can be divided into two steps: \textbf{reference set determination} and \textbf {content-reference mapping}. 

\textbf{Reference set determination} As mentioned in paper's section 3.4, our target is to search for a reference glyph set which covers the most conspicuous-level components, and our detailed algorithm can be demonstrated by Algorithm \ref{alg:refss}. In our algorithm, $X$ denotes the entire character list sorted by glyphs' occurrence frequency, whose total capacity is 20K. $T$ denotes the single level decomposition look-up tables for all characters in $X$. For one key $x_i$ in $T$, their corresponding value $\chi_i$ is a subset of $X$, consisting of their decomposing components' name list and corresponding decomposing form. By 
exploring this components list, we employ breadth first search to recursively obtain all the components at different levels, which forming the component tree in our work. $C$ denotes the set that contains all conspicuous-level components, whose total capacity is \textbf{374}. $N_{ref}$ denotes the total capacity of our target reference set, which is is fixed to \textbf{100} in our experiments. 

As an initialization, we set our target reference set $\hat{U}$ and corresponding covering components $\hat{C}$ as $\emptyset$. Then we begin searching process in $X$. For every glyph $x_i$ in $X$, we obtain its conspicuous-level components $c_i$ via the function \emph{searchComponents} in \ref{alg:refss}. If there is unique components in $c_i$ which is not in current component pool $\hat{C}$, we regard this glyph $x_i$ as our target reference glyph, and we add it and its unique components to $\hat{U}$ and $\hat{C}$ respectively. In our experiment, we ensure that every latest adding reference have \textbf{2} or more unique components. Once the capacity of $\hat{U}$ reaches $N_{ref}$, we terminate the searching process.

\begin{algorithm}
\SetKwInOut{Input}{input}
\SetKwInOut{Output}{output}
\SetKwRepeat{Do}{do}{while}
\Input{$X=\{x_i\}$: Common-used glyph list. \newline 
$T=\{x_i:\chi_i\}$: Single-level decomposition look-up table. $\chi_i \subset X$. \newline
$C$: Conspicuous-level component set. \newline
${N}_{ref}$: Max capacity of Reference set. \newline
$d$: Max search depth in component tree.
}
\Output{$\hat{U}$: Reference set. \newline
$\hat{C}$: All components which $U$ can cover.
}
\SetKwFunction{FMain}{searchComponents}
\SetKwProg{Fn}{Function}{:}{}
\Fn{\FMain{$x_i$}}{
    $queue \leftarrow \{x_i\}$; $i \leftarrow 0$; ${c_i} \leftarrow \emptyset $ \newline
    \Repeat{$i \geq d$}{
      $queue^{*} \leftarrow \emptyset $ \newline
      \For{$x\in queue$}{
        $queue^{*} \leftarrow queue^{*} \cup T(x) $ \newline
        ${c_i} \leftarrow T(x) \cap C $
      }
      $queue \leftarrow queue^{*}$; $i \leftarrow i+1$
    }
    \textbf{return} ${c_i}$ \newline
}
\textbf{End Function}
\newline

\SetKwFunction{FMain}{Main}
\SetKwProg{Fn}{Function}{:}{}
\Fn{\FMain{}}{
    $\hat{U} \leftarrow \emptyset $; $\hat{C} \leftarrow \emptyset $ \newline
    \For{$x_i\in X$}{
        $c_i \leftarrow $ searchComponents($x_i$) \newline
        \lIf{$\exists \hat{c} \in {c}_{i}$ s.t. $\hat{c} \notin \hat{C}$}{
            $\hat{C} \leftarrow \hat{C} \cup {c}_{i}$, add $x_i$ to $\hat{U}$
        }
        \lIf{Len($\hat{U}$) $\geq$ ${N}_{ref}$}{\textbf{break}}
    }
    \textbf{return} $\hat{U}$, $\hat{C} $ \newline
}
\textbf{End Function}
\caption{Reference set selection}
\label{alg:refss}
\end{algorithm}

\textbf {Content-reference mapping} Since we have determined the reference set. It's more convenient for us to find the mapping relationship between a arbitrary glyph and its references whose conspicuous components are in common. For $k$-shot reference mapping (one content glyph corresponds to $k$ style references), we do the following searching process for $k$ times: We find the glyph in reference set which share the most conspicuous component, and remove it from the original reference set. If there is multiple choice, we leave the one whose component's composition form is the same with the content glyph.

We show some of the finding mappings between content and references in Fig. \ref{more results}.

\subsection{Model Architecture}
\begin{figure}   
\centering
\includegraphics[width=\linewidth,scale=1.0]{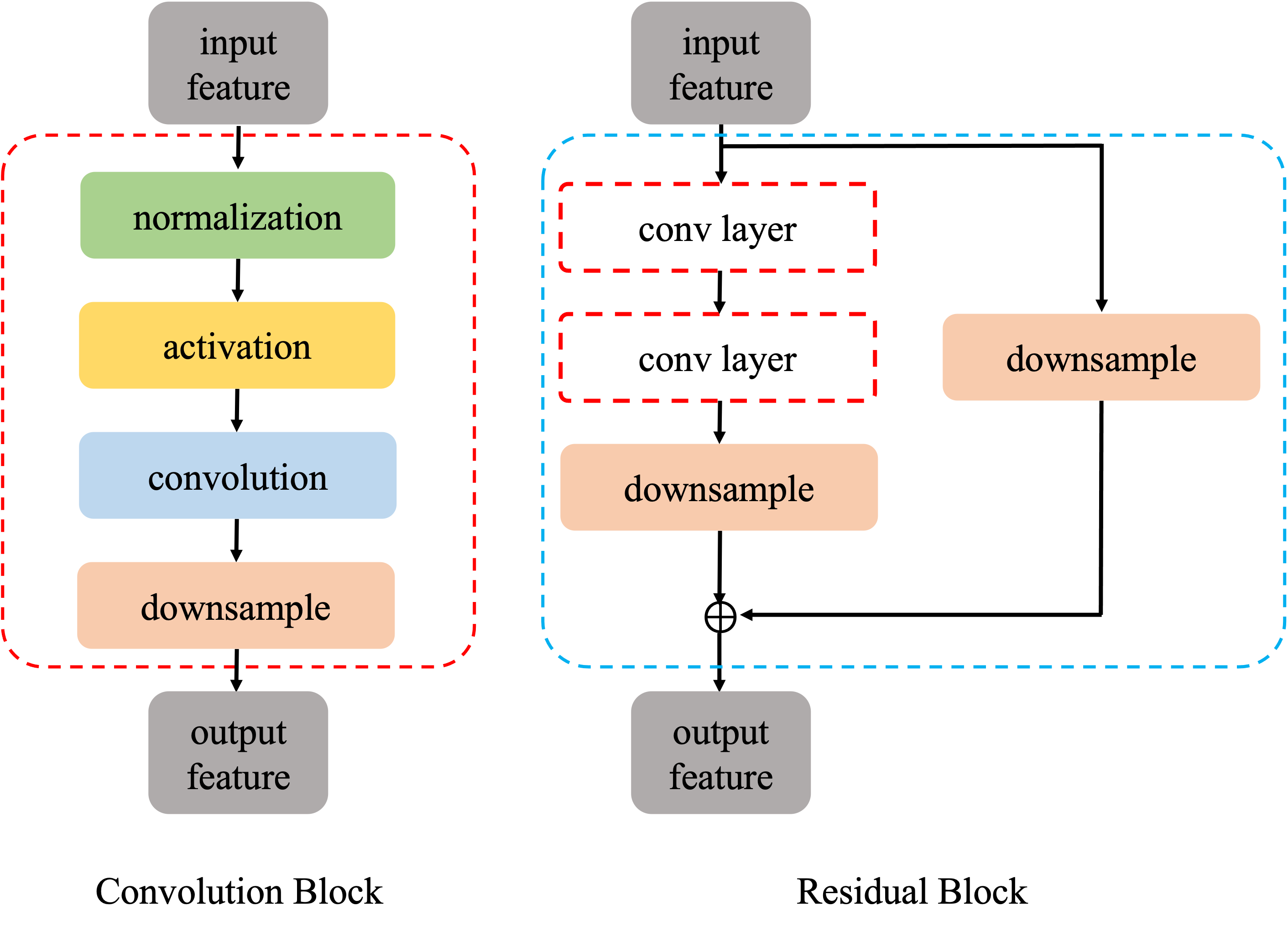}
\caption{Structures of the Convolution Block and the Residual Block in Tab.\ref{architecture}}
\label{fig:blocks}
\end{figure}

Our entire model can be divided into two parts: the generator and discriminator. Both of them are built up of two typical types of blocks: the Convolution block and the Residual block, which are illustrated in Fig \ref{fig:blocks}. The Residual block contains two identical Convolution blocks. Since the Residual block has its own downsampling operator, its Convolution blocks skip the downsampling step. 

\textbf{Generator} As mentioned in our method, the generator model consists of three respective modules: reference encoder $E_r$, content encoder $E_c$, and decoder $D$. Both $E_r$ and $E_c$ are also made up of the following two types of blocks: the Convolution block and Residual block. The detailed architecture is illustrated in Table. \ref{architecture}. 

\textbf{Discriminator} The discriminator is also made up of the Convolution block and the Residual block. In Discriminator, each Convolution and Residual block is followed by a spectral normalization. The detailed architecture is illustrated in Table. \ref{architecture}.

\begin{table*}
  \setlength\tabcolsep{7pt}
  \centering
  \begin{tabular}{l|ccccccc}
      \toprule 
      \multicolumn{8}{c}{\textbf{Reference Encoder $E_r$}} \\ \cmidrule(lr){1-8}
      \cmidrule(lr){1-8}
      Layer Type  & Normalization & Activation & Paddding & Kernel Size & Stride & Downsample & Feature maps \\
      \midrule
      Convolution block  & IN & ReLU & 1 & 3 & 1 & - & 32\\
      Convolution block  & IN & ReLU & 1 & 3 & 1 & AvgPool & 64\\
      Convolution block  & IN & ReLU & 1 & 3 & 1 & AvgPool & 128\\
      Residual block & IN & ReLU & 1 & 3 & 1 & - & 128\\
      Residual block & IN & ReLU & 1 & 3 & 1 & - & 128\\
      Residual block & IN & ReLU & 1 & 3 & 1 & AvgPool & 256\\
      Residual block & IN & ReLU & 1 & 3 & 1 & - & 256\\
      Output Layer & - & Sigmoid & - & - & - & - & 256\\

      \midrule
      
      \midrule
      \multicolumn{8}{c}{\textbf{Content Encoder $E_c$}} \\ \cmidrule(lr){1-8}
      Layer Type  & Normalization & Activation & Paddding & Kernel Size & Stride & Downsample & Feature maps \\
      \midrule
      Convolution block  & IN & ReLU & 1 & 3 & 1 & - & 32\\
      Convolution block  & IN & ReLU & 1 & 3 & 2 & - & 64\\
      Convolution block  & IN & ReLU & 1 & 3 & 2 & - & 128\\
      Convolution block  & IN & ReLU & 1 & 3 & 2 & - & 256\\
      Convolution block  & IN & ReLU & 1 & 3 & 1 & - & 256\\

      \midrule
      
      \midrule
      \multicolumn{8}{c}{\textbf{Decoder $D$}} \\ \cmidrule(lr){1-8}
      Layer Type  & Normalization & Activation & Paddding & Kernel Size & Stride & Upsample & Feature maps \\
      \midrule
      Residual block  & IN & ReLU & 1 & 3 & 1 & - & 256\\
      Residual block  & IN & ReLU & 1 & 3 & 1 & - & 256\\
      Residual block  & IN & ReLU & 1 & 3 & 1 & - & 256\\
      Convolution block  & IN & ReLU & 1 & 3 & 1 & Nearest & 128\\
      Convolution block  & IN & ReLU & 1 & 3 & 1 & Nearest & 64\\
      Convolution block  & IN & ReLU & 1 & 3 & 1 & Nearest & 32\\
      Convolution block  & IN & ReLU & 1 & 3 & 1 & - & 1\\
      Output Layer & - & Sigmoid & - & - & - & - & 1\\

      \midrule
      
      \midrule
      \multicolumn{8}{c}{\textbf{Discriminator $D$}} \\ \cmidrule(lr){1-8}
      \cmidrule(lr){1-8}
      Layer Type  & Normalization & Activation & Paddding & Kernel Size & Stride & Downsample & Feature maps \\
      \midrule
      Convolution block & IN & - & 1 & 3 & 2 & - & 32\\
      Residual block & IN & ReLU & 1 & 3 & 1 & AvgPool & 64\\
      Residual block & IN & ReLU & 1 & 3 & 1 & AvgPool & 128\\
      Residual block & IN & ReLU & 1 & 3 & 1 & AvgPool & 256\\
      Residual block & IN & ReLU & 1 & 3 & 1 & - & 256\\
      Residual block & IN & ReLU & 1 & 3 & 1 & AdaAvgPool & 512\\
      Output layer &       \multicolumn{7}{c}{Multi-task embedding} \\
      \bottomrule 
  \end{tabular}
  \caption{\textbf{Architecture of the generator modules $E_r$, $E_c$, $D$ and the discriminator.} Convolution block and Residual block denote the module mentioned in Fig \ref{fig:blocks}. IN denotes instance normalization. All padding operation is zero-padding. AvgPool and AdaptiveAvgPool denotes average pooling and adaptive average pooling. In discriminator's output layer, we use two embedding operators to embed the output feature map into two prediction vector of the font style and the character's name.}
  \label{architecture}
\end{table*}

\subsection{Training details}
We use Adam optimizer to optimize the FSFont's parameters. The generator and discriminator's learning rates are 0.0002 and 0.0008, respectively. Kaiming initialization is applied for the model. During training, by default we set the reference to be $3$-shot for training. If one glyph's reference set's capacity is less than 3, we duplicate one of reference glyph until the capacity equals to 3 for purpose of batch training. In our proposed SAM, empirically we set number of attention heads to be 8 and batch size to be 32. We train our model with full objective function for 500k iterations. 

\section{Additional Experimental Results}
\begin{figure*}  
\centering
\includegraphics[width=\linewidth,scale=1.00]{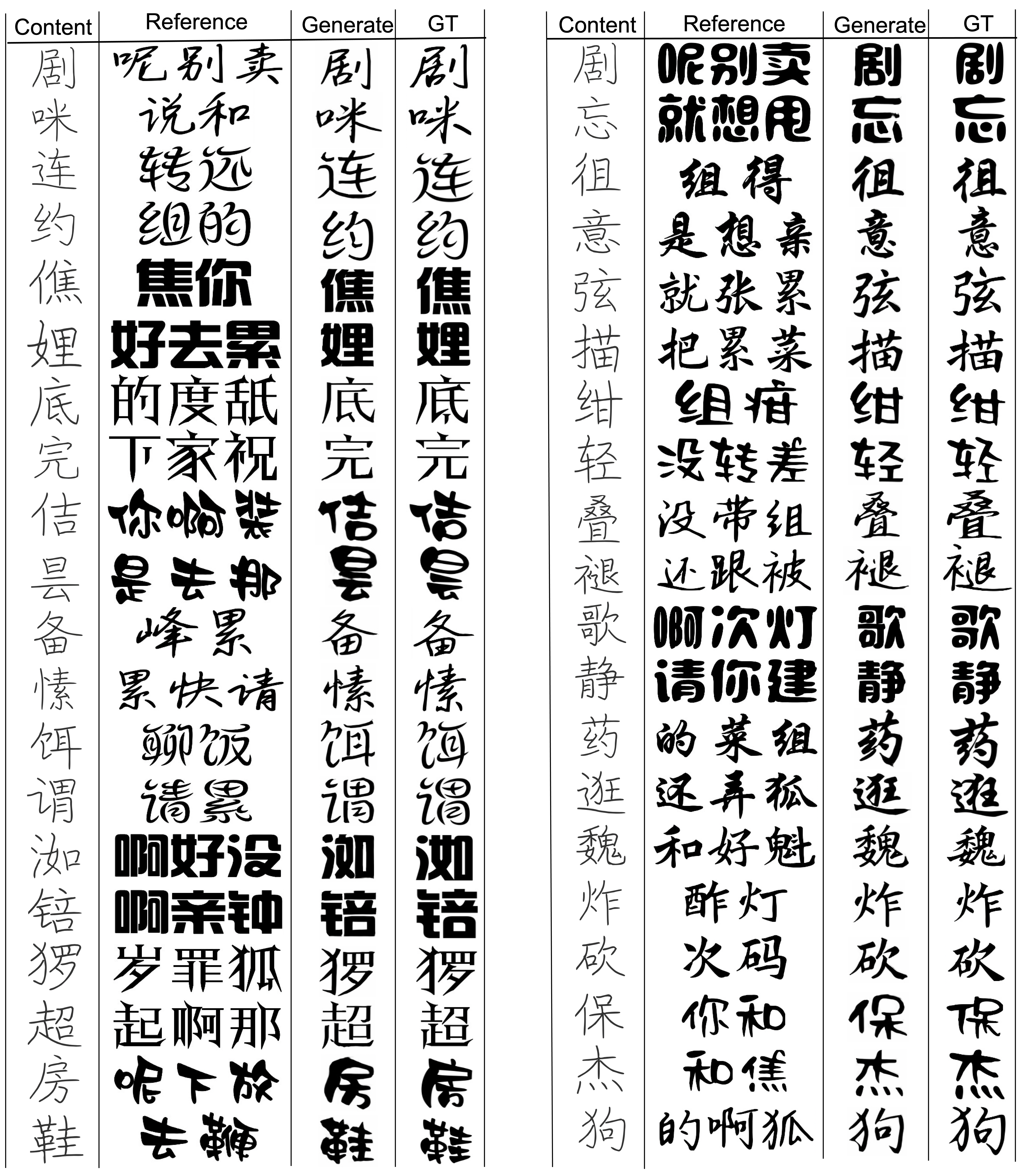}
\caption{\textbf{More Results}. This Figure shows the Content-Reference mapping and generated results on all test fonts.}
\label{more results}
\end{figure*}

\begin{figure*}  
\centering
\includegraphics[width=\linewidth,scale=1.00]{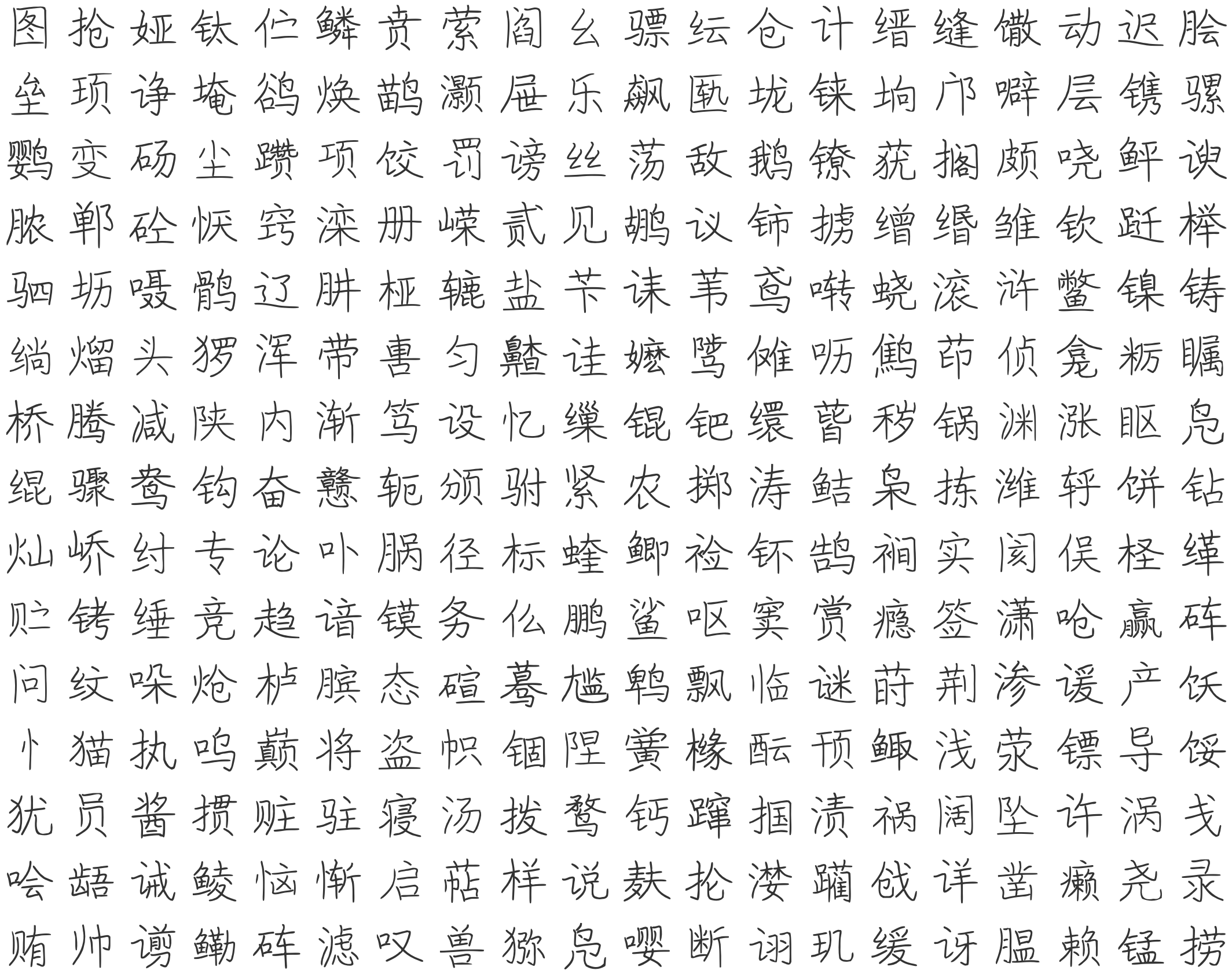}
\caption{\textbf{The content font we used in our experiment.}}
\label{content font}
\end{figure*}

In Fig.\ref{more results}, we demonstrate more experimental results on unseen fonts. The results show that FSFont can deal with variant font styles, including typewriter fonts, artistic fonts, and handwriting fonts. In addition, we show the content font we used in our experiment in Fig.\ref{content font}.

\begin{figure*}  
\centering
\includegraphics[width=\columnwidth,scale=1.0]{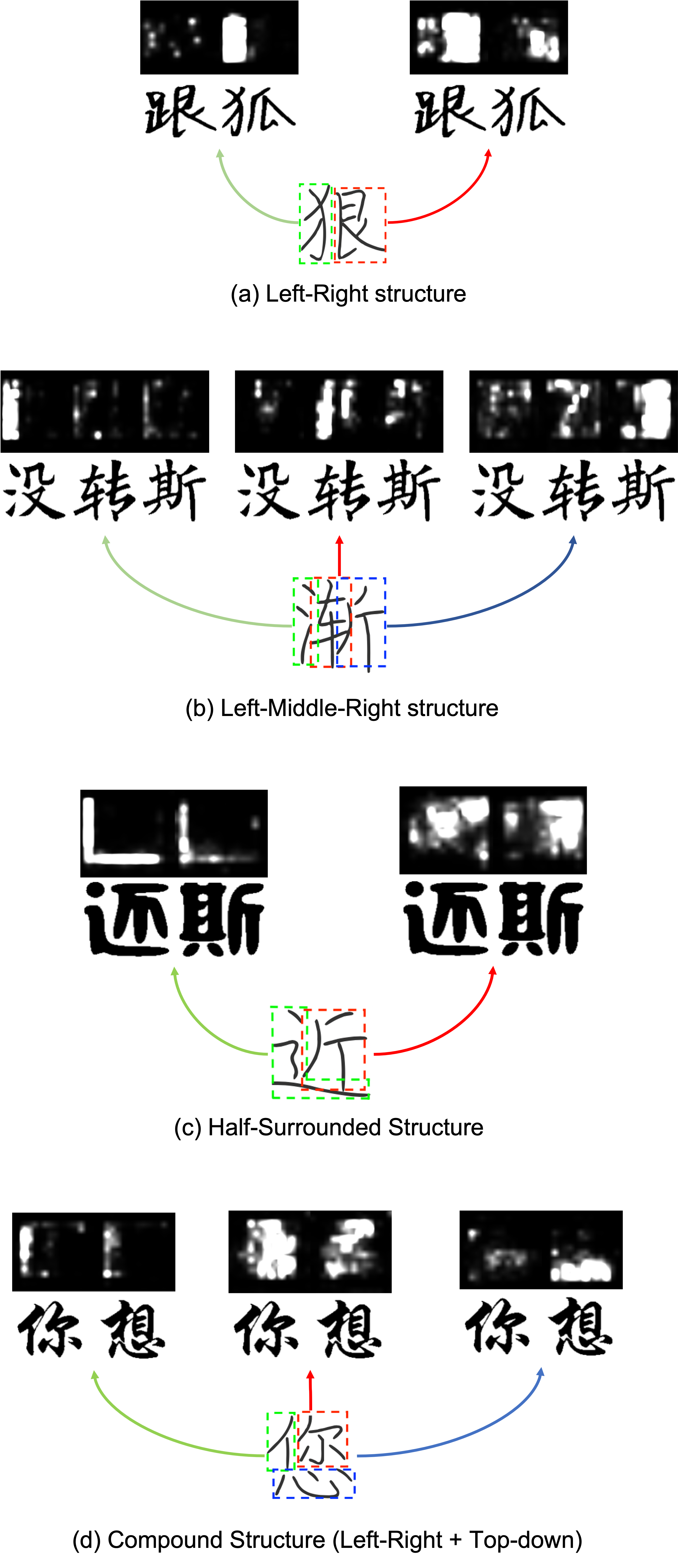}
\caption{\textbf{Visualization on Attention maps}. This Figure shows the attention maps given the different regions in content glyph. }
\label{visual SAM}
\end{figure*}

We also demonstrate more visualization results about the attention maps in Fig. \ref{visual SAM}. We show the attention maps on characters with variant structures. It is shown that the attention is able to capture the correct correspondence in different structures. 

\section{User Study Examples}
\begin{figure*}  
\centering
\includegraphics[width=\linewidth, scale=1.00]{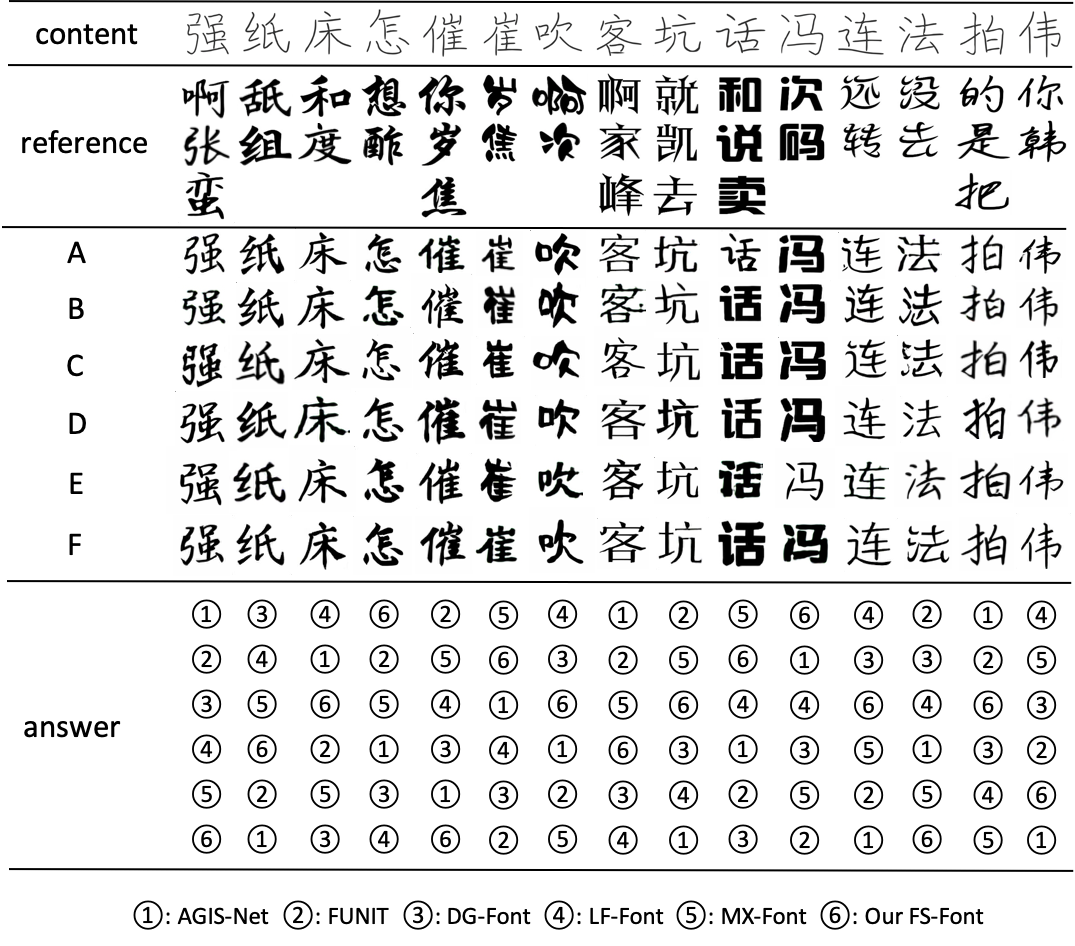}
\caption{\textbf{User studies on UFSC (unseen font seen character) data}. This Figure shows how do we conduct the user study.}
\label{fig:user_study1}
\end{figure*}

\begin{figure*}   
\centering
\includegraphics[width=\linewidth, scale=1.00]{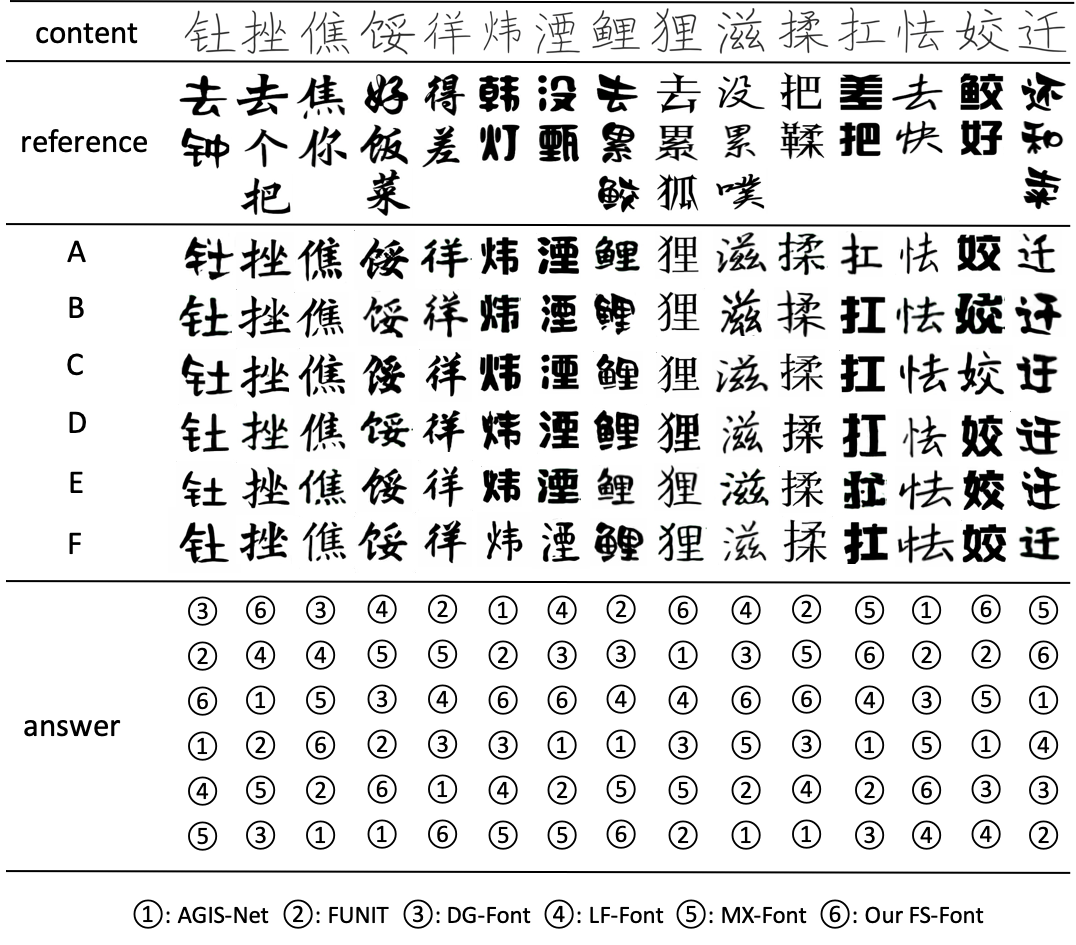}
\caption{\textbf{User studies on UFUC (unseen font unseen character) data}. This Figure shows how do we conduct the user study.}
\label{fig:user_study2}
\end{figure*}

We show the sample images used for the user study in Fig.\ref{fig:user_study1} and Fig.\ref{fig:user_study2}. A content image, a reference set, and shuffled results from five methods, i.e. FUNIT\cite{liu2019few}, AGISNet\cite{chang2018chinese}, LF-Font\cite{park2020few}, DFFont\cite{xie2021dg}, MX-Font\cite{park2021multiple}, and FSFont, are displayed to users for every query. Users are required to choose the most consistent one according the style of the reference set. As the orders of the methods are shuffled, we also provide the answers for each case. 

\end{document}